\documentclass[conference]{IEEEtran}  

\IEEEoverridecommandlockouts                              

\usepackage{graphicx} 
\usepackage[caption=false,font=footnotesize]{subfig}
\usepackage{amsmath} 
\usepackage{amssymb}  
\usepackage{tikz}
\usepackage{pgfplots}
\pgfplotsset{compat=1.18}
\usepackage{siunitx}
\usepackage{booktabs}
\usepackage{bm}
\usepackage{hyperref}
\usepackage{rotating}
\usepackage{cite}
\usepackage[nolist]{acronym}
\usetikzlibrary{arrows.meta, positioning, shapes.geometric, shapes.symbols,
                 calc, fit, backgrounds, decorations.markings, shadows.blur}
\usepackage{xcolor}

\definecolor{eventblue}{RGB}{41,98,163}
\definecolor{procgreen}{RGB}{39,139,89}
\definecolor{measorange}{RGB}{217,119,39}
\definecolor{ekfpurple}{RGB}{118,55,154}
\definecolor{outputred}{RGB}{186,50,50}
\definecolor{lightbg}{RGB}{245,247,250}
\definecolor{stagegray}{RGB}{230,233,238}

\tikzset{
  baseblock/.style={
    draw, rounded corners=3pt, minimum height=9mm,
    text width=22mm, align=center, font=\footnotesize\sffamily,
    inner sep=3pt, line width=0.6pt,
  },
  eventblock/.style={baseblock, fill=eventblue!12, draw=eventblue!70},
  procblock/.style={baseblock, fill=procgreen!12, draw=procgreen!70},
  measblock/.style={baseblock, fill=measorange!12, draw=measorange!70},
  ekfblock/.style={baseblock, fill=ekfpurple!12, draw=ekfpurple!70},
  outblock/.style={baseblock, fill=outputred!12, draw=outputred!70,
                   font=\footnotesize\sffamily\bfseries},
  wideblock/.style={baseblock, text width=32mm},
  xwideblock/.style={baseblock, text width=42mm},
  ioblock/.style={
    draw, rounded corners=2pt, minimum height=7mm,
    text width=18mm, align=center, font=\scriptsize\sffamily,
    inner sep=2pt, line width=0.5pt, fill=white,
  },
  myarrow/.style={-{Stealth[length=2.5mm, width=1.8mm]},
                  line width=0.55pt, color=#1},
  myarrow/.default={black!70},
  dataarrow/.style={myarrow, densely dashed},
  stagefit/.style={rounded corners=5pt, inner sep=5pt,
                   fill=#1, draw=black!20, line width=0.4pt},
  stagefit/.default={stagegray},
  stagelabel/.style={font=\scriptsize\sffamily\bfseries, text=#1},
  stagelabel/.default={black!60},
}

\newcommand{\p}{\mathbf{p}}

\newcommand{\x}{\mathbf{x}}

\acrodef{UAV}[UAV]{Unmanned Aerial Vehicle}
\acrodef{ICP}[ICP]{Iterative Closest Point}
\acrodef{SVM}[SVM]{Support Vector Machine}
\acrodef{HDBSCAN}[HDBSCAN]{Hierarchical Density-Based Spatial Clustering of Applications with Noise}
\acrodef{CC}[CC]{Connected Components}
\acrodef{FFT}[FFT]{Fast Fourier Transform}
\acrodef{P1E}[P1E]{Perspective-1-Ellipse}
\acrodef{MAE}[MAE]{Mean Absolute Error}
\acrodef{RMSE}[RMSE]{Root Mean Square Error}
\acrodef{MAPE}[MAPE]{Mean Absolute Percentage Error}
\acrodef{RPM}[RPM]{Revolutions Per Minute}
\acrodef{KF}[KF]{Kalman Filter}
\acrodef{RTK-GNSS}[RTK-GNSS]{Real-Time Kinematic Global Navigation Satellite System}
\acrodef{IMU}[IMU]{Inertial Measurement Unit}

\title{\LARGE \bf
Relative State Estimation using Event-Based Propeller Sensing
 }

\author{
\IEEEauthorblockN{
Ravi Kumar Thakur*, Luis Granados Segura, Jan Klivan, Radim Špetlík
\\
Tobiáš Vinklárek, Matouš Vrba, Martin Saska
}

\thanks{Authors are with the Department of Cybernetics,  Faculty of Electrical Engineering, Czech Technical University in Prague, Karlovo Namesti 13, 121 35 Prague 2, Czechia. *Corresponding author: {\tt\small ravi.thakur@cvut.cz}}%
\thanks{This work was funded by the Czech Science Foundation (GAČR) under research project no. 26-22419S}

}

\begin{document}

\maketitle
\thispagestyle{empty}
\pagestyle{empty}

\begin{abstract}
 
Autonomous swarms of multi-\ac{UAV} system requires an accurate and fast relative state estimation. Although monocular frame-based camera methods perform well in ideal conditions, they are slow, suffer scale ambiguity, and often struggle in visually challenging conditions. The advent of event cameras addresses these challenging tasks by providing low latency, high dynamic range, and microsecond-level temporal resolution.  This paper proposes a framework for relative state estimation for quadrotors using event-based propeller sensing. The propellers in the event stream are tracked by detection to extract the region-of-interests. The event streams in these regions are processed in temporal chunks to estimate per-propeller frequencies. These frequency measurements drive a kinematic state estimation module as a thrust input, while camera-derived position measurements provide the update step. Additionally, we use geometric primitives derived from event streams to estimate the orientation of the quadrotor by fitting an ellipse over a propeller and backprojecting it to recover body-frame tilt-axis. The existing event-based approaches for quadrotor state estimation use the propeller frequency in simulated flight sequences. Our approach estimates the propeller frequency under $3\%$ error on a test dataset of five real-world outdoor flight sequences, providing a method for decentralized relative localization for multi-robot systems using event cameras.



\end{abstract}


\section{INTRODUCTION}

Autonomous formation flight of a multi-robot aerial swarms requires that robots estimate the states of others accurately. In an aerial swarm, the robots can interact with each other to pursue a common goal. Accurate estimation of linear and rotational kinematic states is necessary to enable multi-robot coordination by performing collision avoidance, infer intent, predicting trajectories and maintaining relative poses with respect to each other. This is commonly achieved by explicit communication between them\cite{chung2018survey} or by using relative localization through various sensing modalities~\cite{walter2019uvdar, ulrich2023real, Faessler14icra}.  Most formation flights involving quadrotors rely on robust communication systems to exchange state information. However, performance can be degraded in the presence of communication latency or packet loss, and in some cases, the stability of formation may be compromised. To solve this problem, we propose event-based propeller sensing to estimate the relative states of quadrotors, thus enabling  decentralized swarm flight in environmentally challenging conditions

\begin{figure}[t]
    \centering
    \includegraphics[width=0.8\columnwidth]{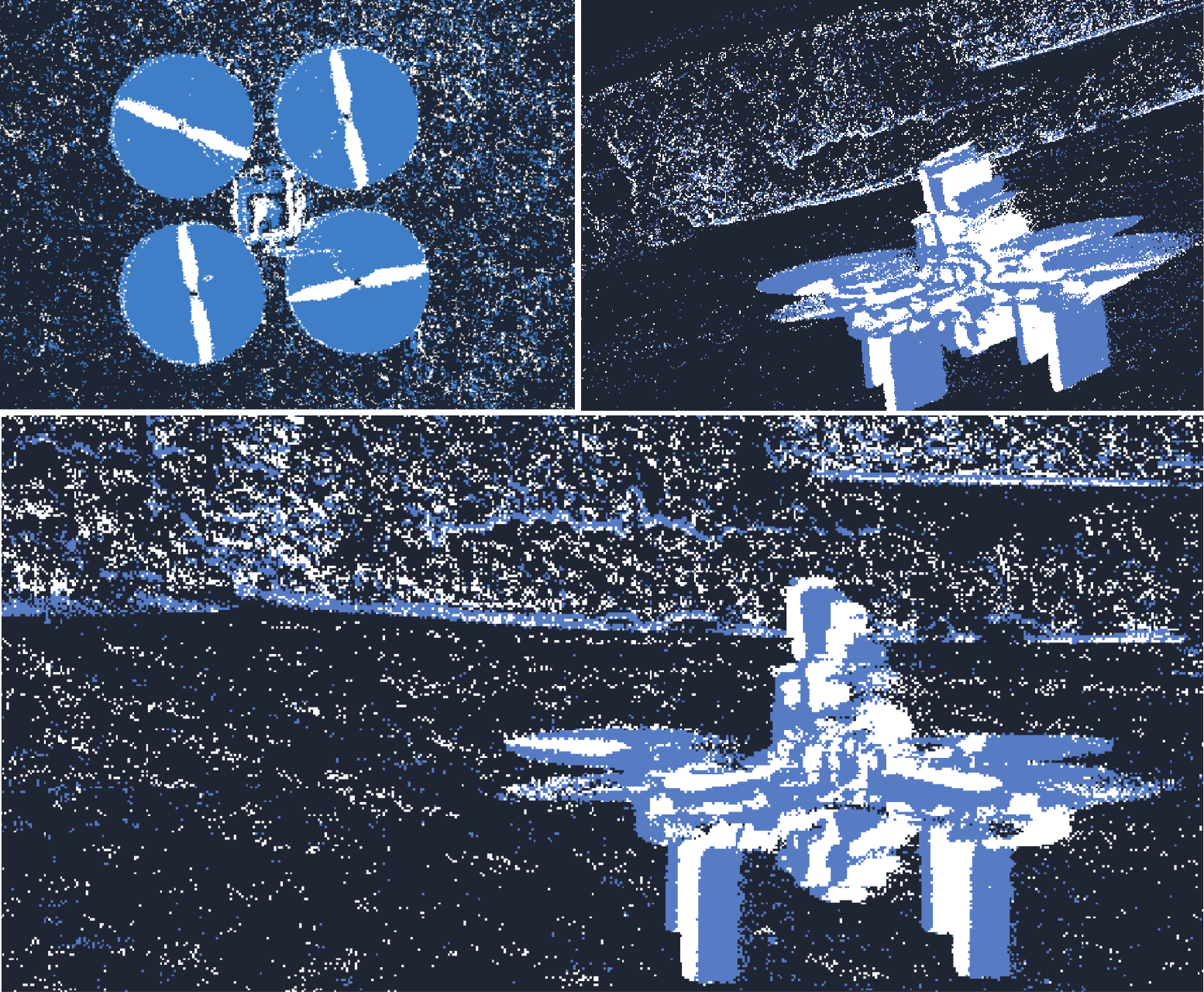}
    \caption{Visuals of a quadrotor in-flight as seen in pseudo-frame representation of streams accumulated from the event camera.}
    \label{fig:intro}
\end{figure}

For an autonomous swarm flight to be resilient it needs the quadrotors to interact locally with limited communication, typically based on vision-based perception. However, vision-based methods rely on a frame-based RGB camera which records the light intensity per pixel for every frame. The frame-based camera has low temporal resolution, high latency, and low dynamic range in comparison to event cameras. These limitations make it a less suitable choice for vision-based state estimation where the environment introduces motion blur, extreme illumination changes, and background noise due to clutter. Thus, frame-based cameras introduce limitations for multi-robot formation flight of autonomous aerial swarms in a perceptually challenging environment.

To address the limitations of frame-based cameras in multi-robot formation flight of aerial swarms, our work relies on event cameras. These sensors are low-powered and operate at high frequency~\cite{gallego2020event}. The sensor area is activated only when there is a change in the visual appearance of the scene. The event streams are generated by any motion in the sensor area. Due to these advantages, event cameras are considered for autonomous navigation tasks where there are dynamic obstacles and the quadrotors are moving at high speed~\cite{vidal2018ultimate, sanket2021evpropnet}. 

Such an autonomous swarm quadrotor system has many applications. It can provide support for  humanitarian assistance and disaster relief missions, survey a large area, and gather spatial intelligence. It can enable planetary exploration with less dependence on the orbiter spacecraft. This system is robust against individual failure and provides efficient spatial coverage for exploration. It can reduce dependency on the orbiter spacecraft for navigation. The distributed network of flying robots can act as a synthetic aperture or form a sensor network~\cite{saska2019large}. Heterogeneous robot teams are proposed for the exploration of canyon-like structures where communication is challenging~\cite{stadler2025exploring, riegler2025towards}.

In this paper, we propose the use of a neuromorphic camera as the primary sensor to perform relative state estimation. Our approach utilizes the event streams from the camera to detect the quadrotor, localize it, and estimate the frequency of its propellers. The frequency of the propeller is used by the state estimation pipeline to get first-order kinematics of the quadrotor. We also make use of the shape of the propeller disk detected in the event streams to estimate the orientation of the quadrotor using \ac{P1E}~\cite{gaudilliere2023perspective}. 

\section{RELATED WORK}

\subsection{Frequency Detection of Periodic Phenomena}

The high dynamic range, low latency and microsecond resolution make event cameras suitable for studying periodic events. They provide a non-invasive way to perform measurements. A kernel-based approach was proposed to estimate angular speeds by manually selecting regions in the event data of a rotating object~\cite{azevedo2025multiple}.  To estimate frequency of any mechanical periodic phenomena like rotation, flicker, vibration etc.~\cite{kolavr2025eeppr} proposed generation of correlation peaks with automatic template selection. In~\cite{zhao2023ev} a combined heatmap and clustering based approach is proposed for propeller detection and subsequently frequency estimation using \ac{ICP} registration. \cite{pfrommer2022frequency} proposed the use of digital infinite impulse response filter for brightness reconstruction and created a per-pixel frequency map by detecting zero-level crossing. A method for creating frequency maps is proposed by applying the \ac{FFT} to the event stream of each pixel\cite{aitsam2024vibration}. These approaches assume a completely periodic motion without any drift or ego motion, thus limiting their application in formation flight. 

\subsection{State Estimation using Event Camera}

The estimation of state of \ac{UAV} is of large interest in swarm robotics. Using an RGB camera for estimation of the relative pose of the camera is a common approach. However, a similar approach cannot be applied to the event camera. The potential use of propeller sensing to estimate rotational attitude parameters was proposed in~\cite{author2024efficient}. In this approach, the event streams are treated as the Poisson process to distinguish events generated by a propeller and the background. The pose is estimated by fitting an ellipse over the aggregated propeller events. HelixTrack extends this work by estimating the pose of the roating propeller using a batched Gauss-Newton optimization. They test their algorithm on propeller dataset with different levels of ego-motion~\cite{helixtrack}. Eventpro~\cite{chen2025count} proposes the estimation of the propeller frequency using hierarchical event processing and known propeller geometry. The flight commands are then predicted using a trained \ac{SVM} classifier. In~\cite{lu2025event}, an estimation of ego-motion of kinematic states based on monocular event vision is proposed, relying on event-based optical flow without propeller detection. Unlike these methods, our approach directly links quadrotors propeller frequencies to a physical dynamics model for state estimation.

\section{STATE ESTIMATION USING PROPELLER SENSING}

\subsection{Event Camera Data Representation}
An event camera responds to any change in brightness in the sensor region. They are low-powered device with high dynamic range and microsecond resolution. In contrast to frame-based cameras, they output asynchronous event streams where each event is represented as a tuple
\begin{equation}
    e_i = (x_i, y_i, t_i, p_i),
\end{equation}
where $(x_i, y_i)$ denote the pixel coordinates, $t_i$ is the microsecond-precision timestamp, and $p_i \in \{-1, +1\}$ indicates the polarity (brightness increase or decrease). An event is triggered when the logarithmic intensity change at a pixel exceeds a threshold $C$:
\begin{equation}
    \left| \log I(x,y,t_i) - \log I(x,y,t_{i-1}) \right| > C.
\end{equation}

This enables the event cameras to respond to fast changes. They naturally respond to features like edges and corners by design. However, because of their inherent sensitivity to motion, they capture background noise as well. The event streams can be visualized using intermediate pseudo-frame representations like time-surface, event histogram and event frame etc. A comprehensive survey of event-based vision, the type of data representation, and their processing is provided in~\cite{gallego2020event} .

\subsection{Propeller Detection Pipeline}
The proposed approach processes event streams in temporal chunks of $\Delta t =$ \SI{10}{\milli\second}. The outdoor data set contains noise generated by the background due to events generated by camera movement. These events are filtered out using spatio-temporal contrast filtering. We implemented two approaches to detect the propeller region.  The first uses \ac{CC} labeling on the event frame and the other uses \ac{HDBSCAN} clustering~\cite{campello2013density} on the event streams. The goal of both detection approaches is to identify the regions of interest for estimating frequency. Detected regions are tracked across frames and assigned stable tracking IDs based on spatial position.

\begin{figure*}[!ht]
\centering
\resizebox{\textwidth}{!}{%
\begin{tikzpicture}[x=1cm, y=1cm,
  B/.style={draw, rounded corners=3pt, minimum height=10mm,
            text width=30mm, align=center,
            font=\footnotesize\sffamily, inner sep=4pt, line width=0.6pt},
  Brpm/.style={B, fill=measorange!12, draw=measorange!70},
  Bcam/.style={B, fill=eventblue!12,  draw=eventblue!70},
  Bobs/.style={B, fill=procgreen!12,  draw=procgreen!70},
  Bori/.style={B, fill=ekfpurple!12,  draw=ekfpurple!70},
  Bout/.style={B, fill=outputred!12, draw=outputred!70,
               font=\footnotesize\sffamily\bfseries, text width=28mm},
  Bio/.style={B, fill=white, draw=black!40, text width=28mm,
              minimum height=11mm},
  Bkf/.style={draw, rounded corners=5pt, align=center,
              font=\footnotesize\sffamily, inner sep=7pt,
              line width=0.9pt},
  arr/.style={-{Stealth[length=2.5mm,width=1.8mm]},
              line width=0.55pt, color=#1},
  arr/.default={black!70},
  couple/.style={-{Stealth[length=3mm,width=2mm]},
                 line width=1.0pt, color=ekfpurple!70, densely dashed}]

  \node[Bio, draw=measorange!60] (rpmin) at (0, 0)
    {RPM$_1$\,--\,RPM$_4$\\[-1pt]{}};

  \node[Bio, draw=ekfpurple!60] (ellin) at (0, -3.0)
    {P1E ellipse\\[-1pt]
     {\tiny best-prop event mask}};

  \node[Bio, draw=eventblue!60] (pixin) at (0, -6.0)
    {Propeller detection\\[-1pt]
     {\tiny centre + mask separation}};

  \begin{scope}[on background layer]
    \node[stagefit=black!4, inner sep=8pt,
          fit=(rpmin)(ellin)(pixin),
          label={[stagelabel=black!90]above:Inputs}] {};
  \end{scope}


  \node[Brpm, text width=32mm] (thrust) at (5.5, 0.55)
    {Total thrust $T$\\[-2pt]
     {\tiny sum of squared motor speeds}};

  \node[Brpm, text width=32mm] (diffrpm) at (5.5, -0.85)
    {Roll differential $\Delta_\mathrm{roll}$\\[-2pt]
     {\tiny right--left motor imbalance}};

  \draw[arr=measorange!70] (rpmin.east) -- ++(8mm,0) |- (thrust.west);
  \draw[arr=measorange!70] (rpmin.east) -- ++(8mm,0) |- (diffrpm.west);

  \begin{scope}[on background layer]
    \node[stagefit=measorange!8, inner sep=7pt,
          fit=(thrust)(diffrpm),
          label={[stagelabel=measorange!90]above:RPM Processing}] {};
  \end{scope}

  \node[Bori, text width=32mm] (backproj) at (5.5, -3.0)
    {Ellipse backproj.\\[-2pt]
     {\tiny disc normal $\hat{\mathbf{b}}_z^\mathrm{raw}$}\\[-1pt]
     {\tiny via conic $\to$ quadric decomp.}};

  \draw[arr=ekfpurple!70] (ellin)--(backproj);

  \begin{scope}[on background layer]
    \node[stagefit=ekfpurple!8, inner sep=7pt,
          fit=(backproj),
          label={[stagelabel=ekfpurple!90]above:P1E Processing}] {};
  \end{scope}

  \node[Bcam, text width=32mm] (depth) at (5.5, -5.0)
    {Depth $Z$ from prop span\\[-2pt]
     {\tiny known size $d$ + focal length $f$}};

  \node[Bcam, text width=32mm] (pinhole) at (5.5, -6.4)
    {Pinhole back-proj.\\[-2pt]
     {\tiny pixel $\to$ 3-D point $\mathbf{p}_c$}};

  \node[Bobs, text width=32mm] (world) at (5.5, -7.8)
    {Cam$\to$FCU$\to$World\\[-2pt]
     {\tiny $\mathbf{p}^W$ via extrinsics + observer pose}};

  \draw[arr=eventblue!70] (pixin.east) -- ++(8mm,0) |- (depth.west);
  \draw[arr=eventblue!70] (depth)--(pinhole);
  \draw[arr=eventblue!70] (pinhole)--(world);

  \begin{scope}[on background layer]
    \node[stagefit=eventblue!8, inner sep=7pt,
          fit=(depth)(pinhole)(world),
          label={[stagelabel=eventblue!90]above:
                 Camera $\to$ World}] {};
  \end{scope}

  \node[Bkf, fill=ekfpurple!6, draw=ekfpurple!60,
        text width=66mm, minimum height=32mm] (orikf) at (12, -1.2)
    {\textbf{Orientation KF}\hfill
     {\small $\hat{\mathbf{b}}_z \in \mathbb{R}^3$\;\;(3 states)}
     \\[6pt]
     \begin{tabular}{@{}p{30mm}@{\hspace{3mm}}p{30mm}@{}}
       \textbf{Predict} & \textbf{Update} \\[2pt]
       {\scriptsize Rotate $\hat{\mathbf{b}}_z$ by}
       &
       {\scriptsize Ellipse-derived} \\[-1pt]
       {\scriptsize RPM angular velocity $\omega_x$}
       &
       {\scriptsize disc normal $\hat{\mathbf{b}}_z^\mathrm{raw}$} \\[3pt]
       {\scriptsize Re-normalise to}
       &
       {\scriptsize Kalman correction} \\[-1pt]
       {\scriptsize $\|\hat{\mathbf{b}}_z\|=1$}
       &
       {\scriptsize of $\hat{\mathbf{b}}_z$}
     \end{tabular}};

  \begin{scope}[on background layer]
    \node[stagefit=ekfpurple!6, inner sep=8pt,
          fit=(orikf),
          label={[stagelabel=ekfpurple!90, xshift=10mm]above:
                 Orientation Estimation}] (orikfbg) {};
  \end{scope}

  \node[Bkf, fill=procgreen!6, draw=procgreen!60,
        text width=66mm, minimum height=32mm] (poskf) at (12, -6.2)
    {\textbf{Position KF}\hfill
     {\small $\mathbf{x}\!=\![\mathbf{p},\,\mathbf{v}]^\top
      \!\in\!\mathbb{R}^6$\;\;(6 states)}
     \\[6pt]
     \begin{tabular}{@{}p{30mm}@{\hspace{3mm}}p{30mm}@{}}
       \textbf{Predict} & \textbf{Update} \\[2pt]
       {\scriptsize Tilt-corrected thrust}
       &
       {\scriptsize World-frame position} \\[-1pt]
       {\scriptsize $\mathbf{a}\!=\!\tfrac{T}{m}\hat{\mathbf{b}}_z\!+\!\mathbf{g}$}
       &
       {\scriptsize $\mathbf{z}\!=\!\mathbf{p}^W$ from camera} \\[3pt]
       {\scriptsize Constant-acceleration}
       &
       {\scriptsize Kalman correction of} \\[-1pt]
       {\scriptsize double integrator}
       &
       {\scriptsize $\hat{\mathbf{p}}$ and $\hat{\mathbf{v}}$}
     \end{tabular}};

  \pgfmathsetmacro{\thrustX}{7.6}          
  \pgfmathsetmacro{\gapY}{-3.7}            
  \draw[arr=measorange!70, line width=0.8pt]
    let \p1 = ([xshift=-15mm]poskf.north) in
    (thrust.east) -- (\thrustX, 0.55)       
    -- (\thrustX, \gapY)                    
    -- (\x1, \gapY)                         
    -- (\p1)                                
    node[pos=0.20, right=1.5mm, font=\scriptsize\sffamily, text=black!55]
    {$T$};

  \draw[arr=procgreen!70, line width=0.8pt]
    (world.east) -- ++(6mm,0) |- (poskf.west)
    node[pos=0.25, above=1.5mm, font=\scriptsize\sffamily, text=black!55]
    {};

  \begin{scope}[on background layer]
    \node[stagefit=procgreen!6, inner sep=8pt,
          fit=(poskf),
          label={[stagelabel=procgreen!90, xshift=10mm]above:
                 Position Estimation}] (poskfbg) {};
  \end{scope}

  \draw[couple]
    ([xshift=-5mm]orikf.south) -- ([xshift=-5mm]orikf.south |- poskf.north)
    node[midway, right=2mm, font=\scriptsize\sffamily\bfseries,
         text=ekfpurple!80]
    {$\hat{\mathbf{b}}_z$ {\normalfont\tiny (thrust dir.)}};

  \node[Bout] (rpout)  at (18, -1.2)
    {Roll $\hat\phi$,\; Pitch $\hat\theta$};
  \node[Bout] (posout) at (18, -5.4)
    {Position $\hat{\mathbf{p}}$};
  \node[Bout] (velout) at (18, -7.0)
    {Velocity $\hat{\mathbf{v}}$};

  \draw[arr=ekfpurple!70, line width=0.8pt]
    (orikf.east) -- (rpout.west)
    node[midway, above=1.5mm, font=\tiny\sffamily, text=black!50]{};

  \draw[arr=procgreen!70, line width=0.8pt]
    (poskf.east) -- ++(8mm,0) |- (posout.west);
  \draw[arr=procgreen!70, line width=0.8pt]
    (poskf.east) -- ++(8mm,0) |- (velout.west);

    \begin{scope}[on background layer]
    \node[stagefit=outputred!6, inner sep=8pt,
          fit=(posout)(velout)(rpout),
          label={[stagelabel=outputred!90]above:Output Estimate}] {};
  \end{scope}

\end{tikzpicture}
}%
\caption{State estimation pipeline.
  Two coupled \acp{KF} operate on complementary measurements. Orientation \ac{KF} tracks the body $z$-axis
  $\hat{\mathbf{b}}_z$, predicted by \ac{RPM}-derived angular velocity and corrected by \ac{P1E} ellipse backprojection.
  Position \ac{KF} tracks position and velocity, predicted by tilt-corrected thrust and corrected by camera-derived world-frame position. 
  The orientation estimate provides the thrust direction for the position prediction step. Roll and pitch are extracted from $\hat{\mathbf{b}}_z$ via a yaw-free  decomposition since yaw is unobservable from the propeller ellipse.
  }
\label{fig:state_flow}
\end{figure*}
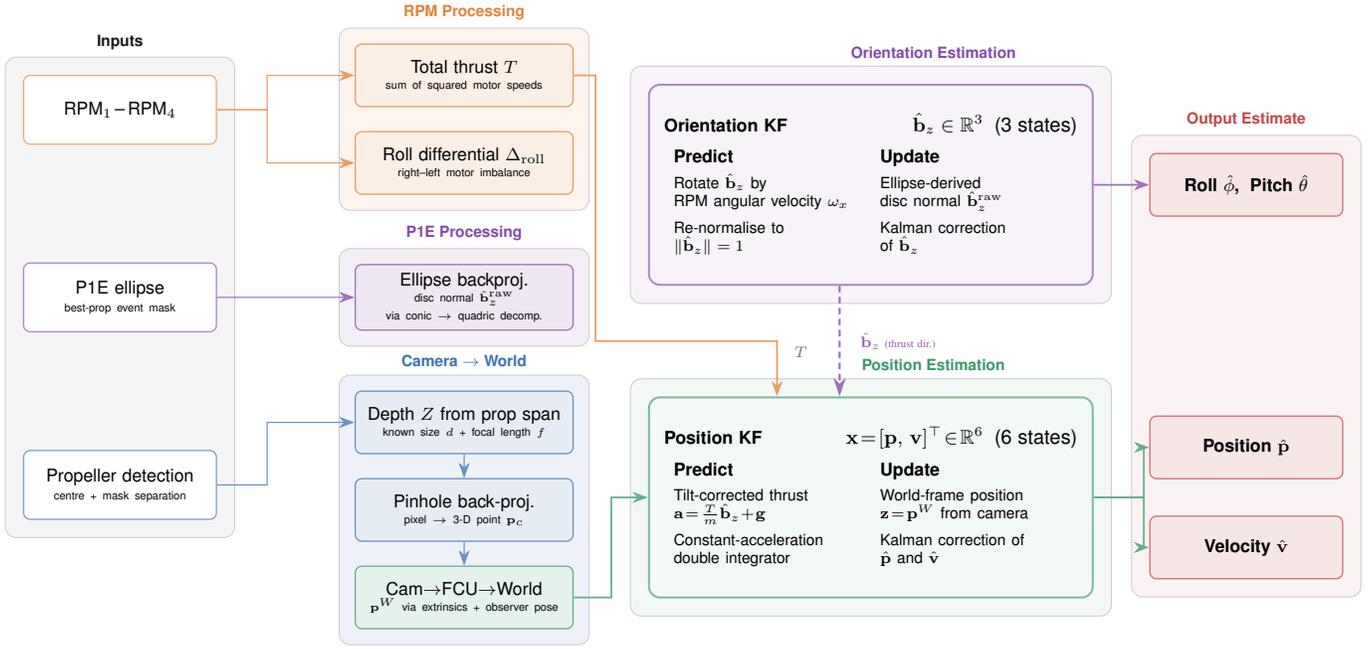


\subsubsection{\acl{CC} Detection}
The \ac{CC} detector spatially accumulates events within each chunk to form a binary event density map. This frame undergoes a morphological erosion to separate merged blob regions to avoid false detection. The \ac{CC} analysis~\cite{samet2002efficient} extracts blob properties. The regions with area below a threshold are rejected as non-propeller region. This approach was found to be computationally efficient but sensitive to parameter tuning.

\subsubsection{\ac{HDBSCAN} Clustering Detection}
The density based clustering approach treats each event as a two-dimensional point and identifies dense spatial clusters without assuming circular geometry. Due to its ability to work directly on the event streams, an intermediate frame-like representation is not required. For computational efficiency, we subsample the number of events. To detect the propellers, parameters such as size, the minimum number of clusters, minimum number of samples are tuned.

For each extracted cluster, we fit an ellipse via principal component analysis of the spatial covariance matrix. The cluster centroid is $\mu_j$ and the $2\times 2$ covariance matrix is:

\begin{equation}
    \boldsymbol{\Sigma}_j = \frac{1}{n_j - 1} \sum_{i=1}^{n_j} (\mathbf{r}_i - \boldsymbol{\mu}_j)(\mathbf{r}_i - \boldsymbol{\mu}_j)^T,
\end{equation}

where, $n_{j}$ is the number of events in $j$th cluster. The position of each event within a cluster is given by $r_{i}$  The eigendecomposition of the covariance matrix yields two eigenvalues and corresponding eigenvectors, which are used to determine the semi-axes of the fitted ellipse. The detected clusters are rejected if the major axis and the number of events are below a certain threshold. The advantage of density-based clustering is its adaptability when propellers have non-circular appearances, specially when seen from oblique viewing angles. Compared to component labeling, this approach is computationally expensive. 

\subsection{Tracking of the quadrotor}
The detected propeller regions were tracked across temporal chunks using Norfair~\cite{joaquin_alori}, an open-source multi-object tracking framework. This tracker maintains track continuity by associating detections between frames using Euclidean distance in position space. These detections can come from either \ac{CC} labeling or density based clustering. Each detection is represented by its centroid $(c_x, c_y)$, which serves as both the tracked point and the re-identification embedding for similarity matching. The four centroids can be used to derive the center of the quadrotor and track its position across time. A general survey of event-based quadrotor detection methods is given in \cite{magrini2025drone}. In this work, the objective of propeller detection and tracking is also to validate the accuracy of frequency estimation by matching the tracked propeller ID with the ground truth. 

The tracker employs a distance threshold and maintains tracks for a few frames during temporary occlusions or detection failures. When a propeller is briefly missed (e.g., when detection fails due to \ac{CC} labeling merging the blobs or occlusion of the propellers), the tracker predicts its position based on motion history, preventing track fragmentation. This temporal consistency is crucial for maintaining continuous frequency estimates throughout the flight sequence.

\subsection{Propeller Frequency Estimation}
The frequency is computed by observing the event counts in the region of interest defined by a sub-quadrant in the bounding box around detected propeller. In these regions, we temporally bin the events and create a sliding window consisting of time bins sufficient to capture multiple blade passages, resulting in an event-count signal. The \ac{FFT} is applied to the binned event count signal to detect the dominant frequency peak corresponding to the blade passage rate. A \ac{KF} smooths the raw estimates. The filter is initialized from the first \ac{FFT} measurement to avoid convergence delays. The  filter provides temporal consistency by rejecting outliers, smoothing out measurement noise.

\begin{figure}[t]
    \centering
    \includegraphics[width=0.8\columnwidth]{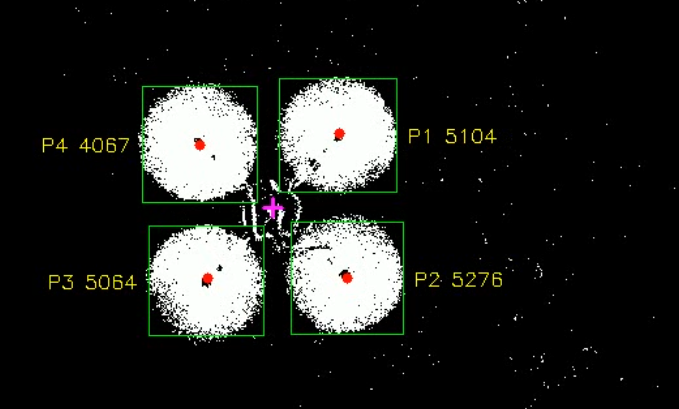}
    \caption{The event frame showing instantaneous \ac{RPM} and center of quadrotor. The propellers are numbered 1--4 from top-right.}
    \label{fig:prop_id}
\end{figure}

\subsection{Relative State Estimation}

The kinematics and dynamics model of the quadrotor is formulated as Eq.~\eqref{eq:quad_kinematics1}--Eq.~\eqref{eq:quad_dynamics2}. The equations describing thrust $T$ is generated along the body $z$-axis proportional to the square of the motor angular velocity $\Omega_{i}$. The rotation matrix corresponding to the quaternion $\mathbf{q}$ is denoted by $\mathbf{R}(\mathbf{q})$. It converts vectors in the body-fixed frame of the quadrotor to the world frame. The total thrust of the target quadrotor is obtained as $T = c_f \sum_{i=1}^{4}\Omega_i^2$, with $\Omega_i$ (\si{\radian\per\second}) denoting the angular velocity of motor $i$ obtained from the event-camera propeller frequency estimates and $c_{f}$ denoting thrust coefficient of the motor. The position and velocity are denoted by $\mathbf{p}$ and $\mathbf{v}$ respectively. A detailed modeling of dynamics and control of a quadrotor system is described in~\cite{mahony2012multirotor}.

\begin{equation}
\dot{\mathbf{p}} = \mathbf{v},
\label{eq:quad_kinematics1}
\end{equation}
\begin{equation}
\dot{\mathbf{v}} = \mathbf{R}(\mathbf{q})
\begin{bmatrix}0\\0\\T/m\end{bmatrix} + \mathbf{g},
\label{eq:quad_dynamics1}
\end{equation}
\begin{equation}
\dot{\mathbf{q}} = \tfrac{1}{2}\,\mathbf{q}\otimes[0,\,\boldsymbol\omega],
\label{eq:quad_kinematics2}
\end{equation}
\begin{equation}
\dot{\boldsymbol\omega} = -\beta\,\boldsymbol\omega,
\label{eq:quad_dynamics2}
\end{equation}

the relative state estimation pipeline has position and orientation estimation modules as shown in Fig.~\ref{fig:state_flow}. The position estimation module has a vector $\mathbf{x}=[\mathbf{p}^\top,\ \mathbf{v}^\top]^\top \in \mathbb{R}^6$. The horizontal components do not receive thrust input and are propagated by constant-velocity extrapolation, making the contribution of propeller frequency directly observable in vertical tracking performance. The state is tracked with a standard linear \ac{KF}. 

The orientation is estimated by the tracking $z$-axis $\hat{\mathbf{b}}_z= [0,0,1]^\top \in \mathbb{R}^3$ representing thrust direction in the world frame. The propeller frequency estimates are fused with camera-derived quadrotor position measurements in the world in this filter. The filters are coupled such that orientation estimate provides the thrust direction used by the position estimation module. The total thrust as a function of the motor constant and the propeller frequencies gives the world frame input acceleration. We describe both the modules below. 

\subsubsection{Position Estimation}
The world-frame acceleration is computed using $\hat{\mathbf{b}}_z$ from the orientation filter. This provides a transformation to obtain the acceleration in the world frame as  
\begin{equation}
\mathbf{a} = \frac{T}{m}\,\hat{\mathbf{b}}_z + \mathbf{g},
\end{equation}
where $m$ is the mass of the quadrotor and gravity is denoted by $\mathbf{g} = [0,0,-9.81]^\top$\,\si{\meter\per\second\squared}. When the orientation estimate is not yet available, the filter defaults to the hover assumption $\hat{\mathbf{b}}_z = [0,0,1]^\top$, so that thrust contributes only to the vertical component. The state transition matrix  $\mathbf{F}$ and input $\mathbf{u}$ gives the prediction model as \eqref{eq:state_propogation} and \eqref{eq:covariance_propagation} \:
\begin{equation}
\mathbf{F} =
\begin{bmatrix}
\mathbf{I} & \mathbf{I}\Delta t \\
\mathbf{0} & \mathbf{I}
\end{bmatrix}, \quad
\mathbf{u} =
\begin{bmatrix}
\tfrac{1}{2}\mathbf{a}\Delta t^2 \\
\mathbf{a}\Delta t
\end{bmatrix},
\end{equation}

\begin{equation}
\hat{\mathbf{x}}_{k+1|k} = \mathbf{F}\hat{\mathbf{x}}_{k|k} + \mathbf{u},
\label{eq:state_propogation}
\end{equation}
\begin{equation}
\mathbf{P}_{k+1|k} = \mathbf{F}\mathbf{P}_{k|k}\mathbf{F}^\top + \mathbf{Q}\Delta t,
\label{eq:covariance_propagation}
\end{equation}

where $\mathbf{P}$ is the state error covariance matrix and $\mathbf{Q}$ is the process noise covariance. 

The world-frame position measurement $\mathbf{z}_k$ is derived from event-frame detections. A known physical diagonal length between the propellers and the pixel length between the detected propellers is used to estimate depth using the stereo disparity and camera intrinsics. The depth obtained is used to convert the coordinate of the quadrotor center to the world frame. The downward-facing event camera is mounted on the observer quadrotor at a known position and orientation. The quadrotor center in the camera frame is transformed to the observer frame using the rigid transformation. Using the known pose of the observer, this quadrotor center is transformed from the observer frame to the world frame. The position  \ac{KF} observation matrix and gain is

\begin{equation}
    \mathbf{H} = [\mathbf{I}_3\ \mathbf{0}_3],
\end{equation}
\begin{equation}
    \mathbf{K} = \mathbf{P}\mathbf{H}^\top(\mathbf{H}\mathbf{P}\mathbf{H}^\top + \mathbf{R})^{-1},
\end{equation}

the monocular depth estimation using the event frame has inherent uncertainty. To address that, we use measurement noise covariance $\mathbf{R} = \mathrm{diag}(\sigma_\mathrm{lat}^2, \sigma_\mathrm{lat}^2, \sigma_\mathrm{depth}^2)$, with separate lateral and depth noise terms.

\subsubsection{Orientation Estimation}

The orientation prediction makes use of differential propeller frequencies which produces a roll angular velocity in the body frame~\cite{mahony2012multirotor} as
\begin{equation}
\omega_x^\mathrm{body} = k_\mathrm{roll}\!\left(\sum_{i \in \text{right}}\!\Omega_i^2 - \sum_{i \in \text{left}}\!\Omega_i^2\right)
\end{equation}
The body-frame angular velocity $\boldsymbol\omega^\mathrm{body}$ is rotated into the world frame using the current $\hat{\mathbf{b}}_z$ and used to propagate the unit vector using a constant angular velocity model
\begin{equation}
\hat{\mathbf{b}}_{z,k+1|k} = \hat{\mathbf{b}}_{z,k} + (\boldsymbol\omega_w \times \hat{\mathbf{b}}_{z,k})\,\Delta t, \quad
\hat{\mathbf{b}}_z = \frac{\hat{\mathbf{b}}_z}{\|\hat{\mathbf{b}}_z\|}
\label{eq:ori_predict}
\end{equation}
with covariance update given as follows
\begin{equation}
 \mathbf{P}_\mathrm{ori} = \mathbf{P}_\mathrm{ori} + \mathbf{Q}_\mathrm{ori}\,\Delta t.   
\end{equation}

The event stream of a  propeller disk forms an ellipse on the image plane of the event frame. We find the best fit ellipse\cite{fitzgibbon1996direct} on the propeller with highest number of events. The ellipse encodes orientation information\cite{gaudilliere2023perspective, lo2002trip}. We estimate the normal of the propeller by reverse projection of the best-fit ellipse through the intrinsics of the camera and the \ac{P1E} formulation~\cite{gaudilliere2023perspective}. This formulation yields a measurement of the body $z$-axis $\hat{\mathbf{b}}_z$ in the world frame. The standard Kalman update is applied with $\mathbf{H} = \mathbf{I}_3$. The measurement noise covariance $\mathbf{R}_\mathrm{ori} = \sigma_m^2\,\mathbf{I}_3$ reflects uncertainty in ellipse backprojection. 

Roll and pitch are extracted from the filtered $\hat{\mathbf{b}}_z$ via a yaw-free decomposition:
\begin{equation}
\phi = \arcsin(-\hat{b}_{z,y}), \quad \theta = \arctan2(\hat{b}_{z,x},\; \hat{b}_{z,z}),
\end{equation}
the yaw component is not observable from the propeller ellipse. When the disk rotates about its normal, it produces an identical projection, so yaw cannot be recovered from the ellipse geometry alone~\cite{gaudilliere2023perspective}. Thus, we restrict the orientation estimation to roll and pitch.

\begin{figure}[tb]
    \centering
    \subfloat[Observer-Target Distance Range per Test Sequence ]{%
        \includegraphics[width=\columnwidth]{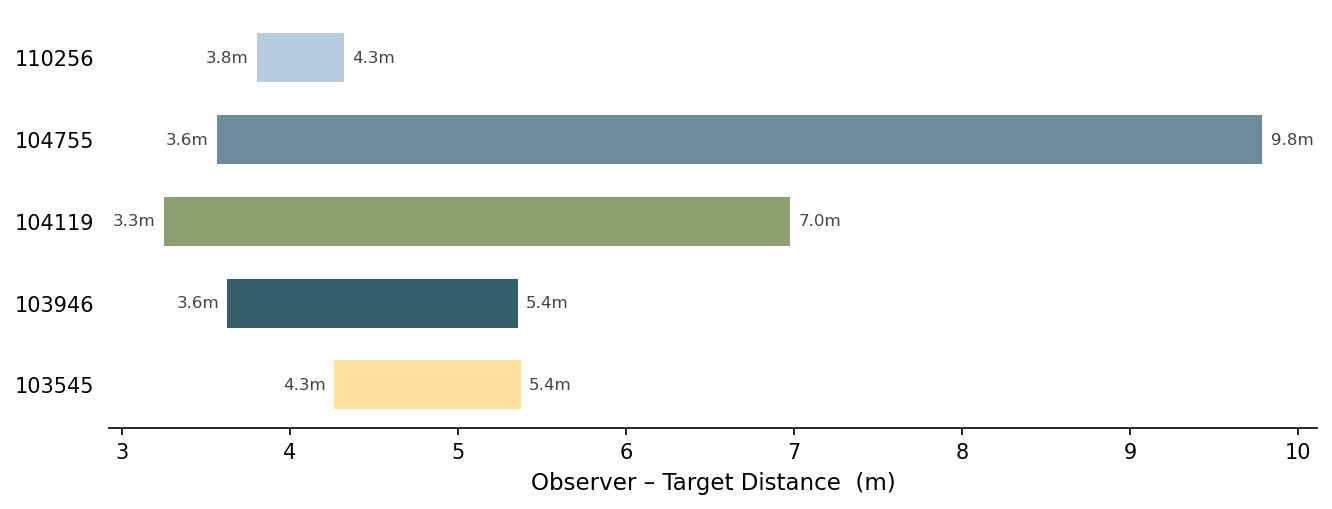}
        \label{fig:range}
    }

    \vspace{0.5cm}

    \subfloat[Detection Recall vs Observer-Target Distance ]{%
        \includegraphics[width=\columnwidth]{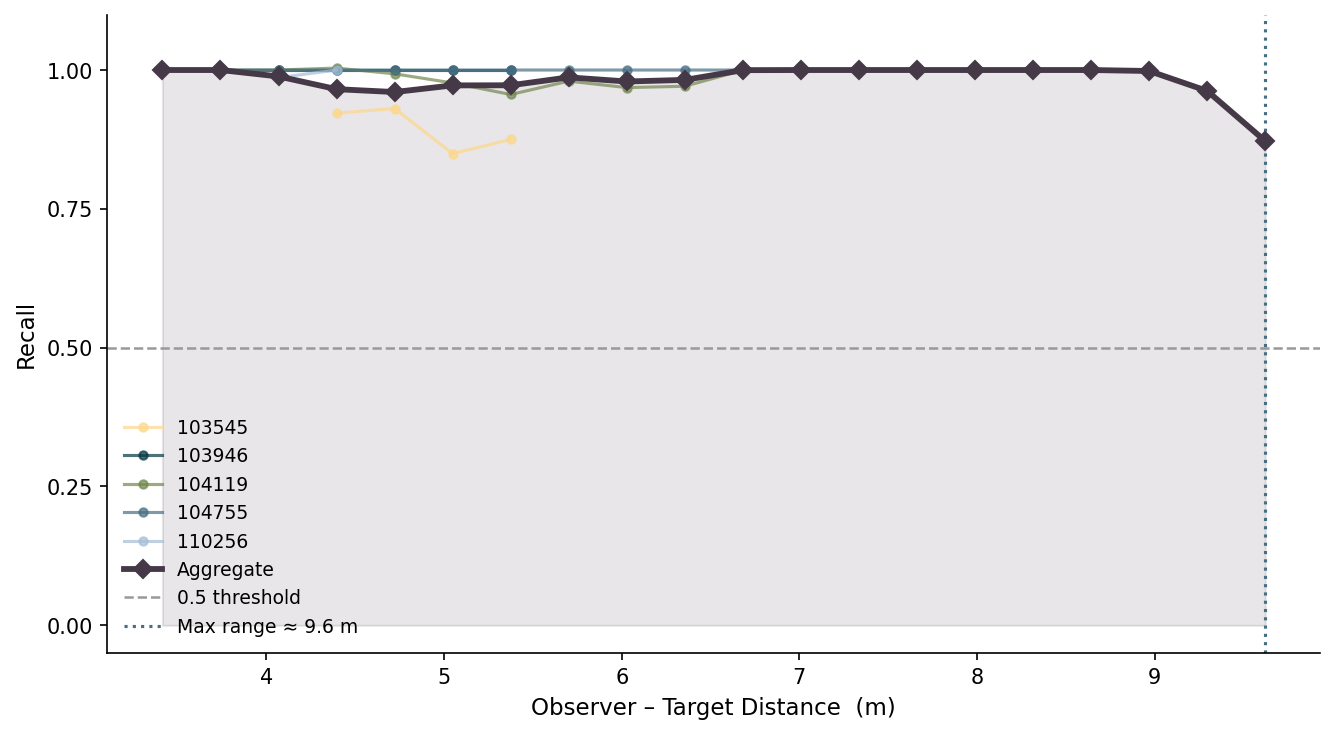}
        \label{fig:recall}
    }

    \caption{The propeller detection results on the test dataset a) The figure shows range of distance between the target and the observer quadrotor in each sequence, b) shows detection recall vs the distance.}
    \label{fig:distance_recall}
\end{figure}

\section{EXPERIMENTS AND RESULTS}

\subsection{Experimental Setup}

The experiments were conducted outdoors using two quadrotors in a leader-follower formation. The follower acted as an observer with a downward-facing event camera. It carried the Prophesee IMX636 event sensor, such that the target's spinning propellers were in the camera's field of view during flight. The leader was designated as the target. A total of six flight sequences were recorded in the outdoor environment. The observer moved at nearly fixed altitude while the target performed a variety of maneuvers beneath it while varying the vertical separation between the two vehicles. In some trials, the target performs aggressive lateral motions. In all the flight sequences, the four propellers of target are visible, except a few instances where it went out the frame. The performance of the propeller detection and the range of distances in the test sequences is shown in Fig.~\ref{fig:distance_recall}

\begin{figure}[tbp]
    \centering
    \includegraphics[width=0.95\linewidth]{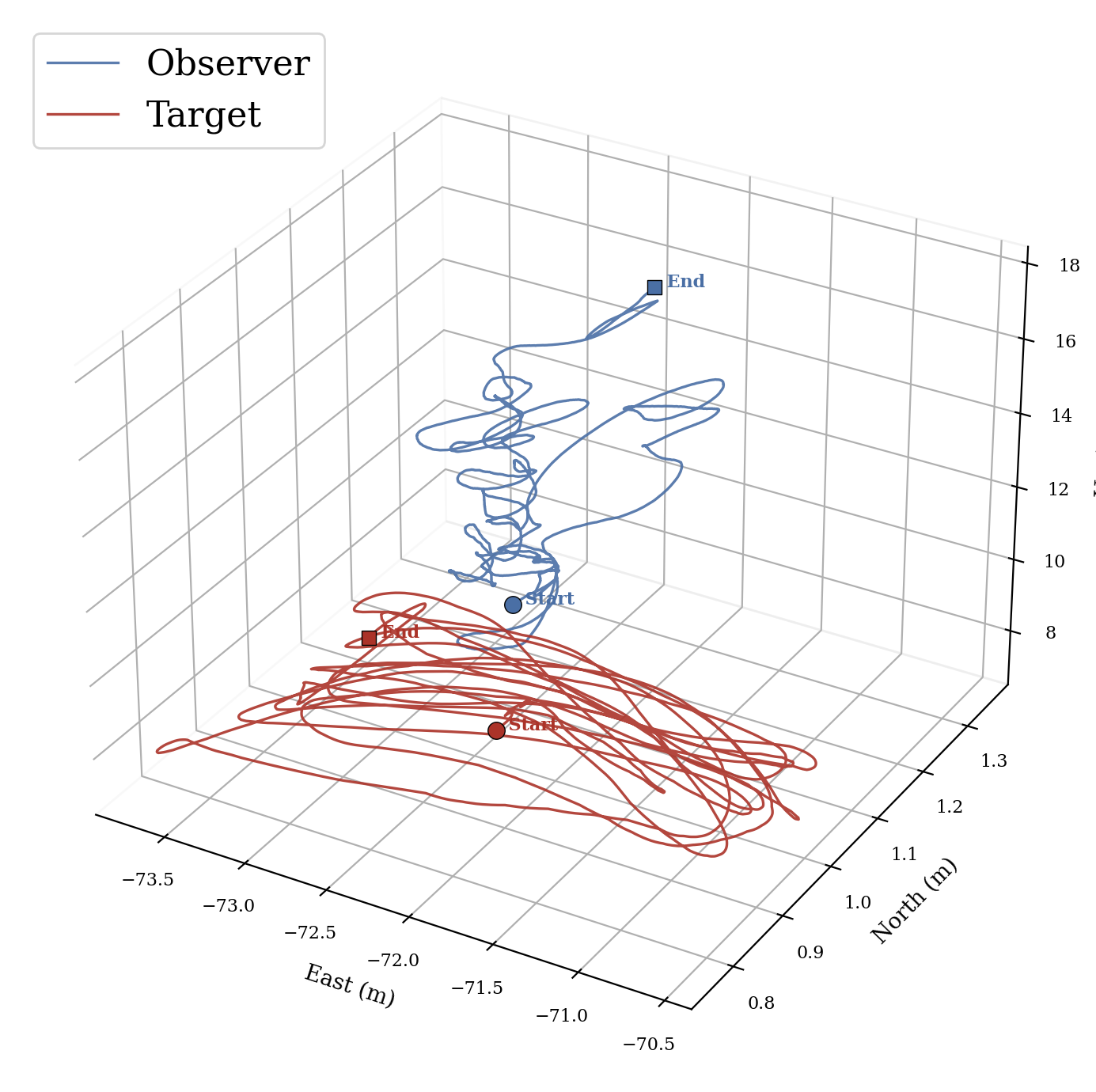}
    \caption{Target-observer 3-D trajectories in a single validation flight sequence}
    \label{fig:trajectory}
\end{figure}

Telemetry was performed using \ac{RTK-GNSS} and \ac{IMU} to record the position and orientation of the two quadrotors. In addition to states, the propeller \ac{RPM} ground truth was obtained using an external sensor and logged synchronously with the event stream. For the purpose of development and parameter tuning, we used one of the sequences as validation data. This validation sequence had the target's four propellers operated between \SI{143}{\hertz} and \SI{197}{\hertz} with a mean of approximately \SI{166}{\hertz}, \ac{RPM} across all motors. The trajectory of the target and the observer in the world frame is shown in Fig.~\ref{fig:trajectory}.

\begin{figure*}[htb]
    \centering
    \subfloat[\ac{RPM} error distribution ]{%
        \includegraphics[width=0.48\textwidth]{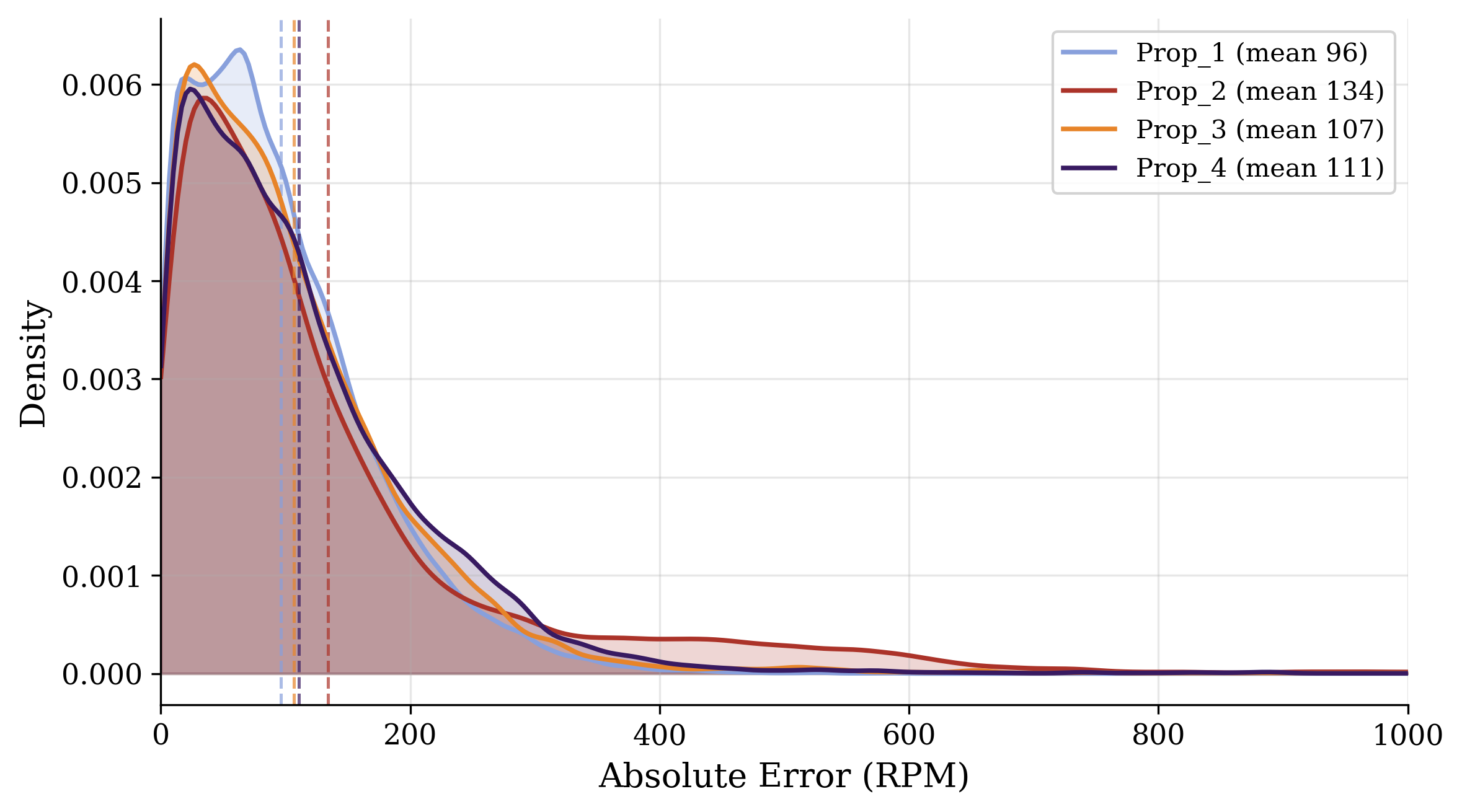}
        \label{fig:cc_hist_1}
    }
    \hfill
    \subfloat[Error over time ]{%
        \includegraphics[width=0.48\textwidth]{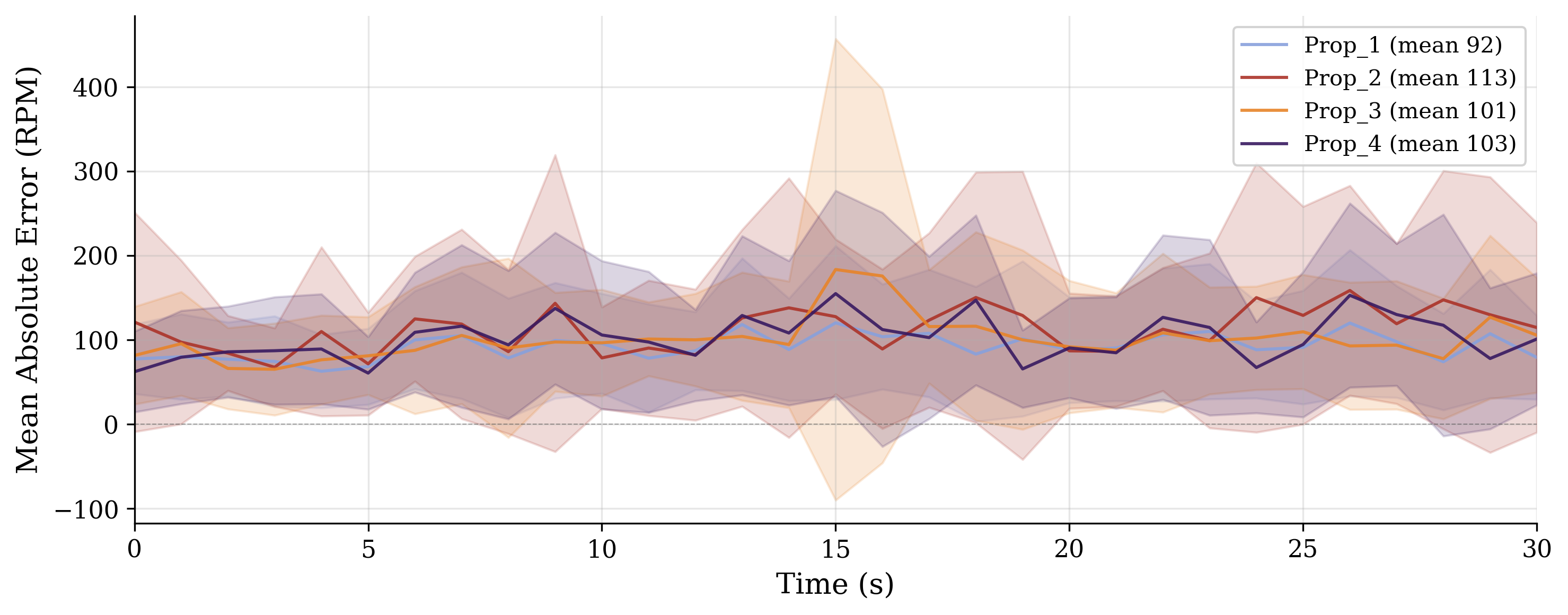}
        \label{fig:hdbscan_hist_1}
    }

    \caption{The \ac{RPM} error analysis for the connected component labeling based propeller detection. a) The figure shows per-propeller absolute error distributions across the five test flight  sequences, b) shows propeller frequency estimation error over time. The error was nearly identical for the \ac{HDBSCAN} based method.}
    \label{fig:error_dist}
\end{figure*}

\subsection{Aggregate \ac{RPM} Results}
For each propeller, we computed \ac{MAE}, \ac{RMSE}, and \ac{MAPE}. This was performed for the five flight sequences in the test data using both the propeller detection approaches, the density based clustering method and connect component labeling. Table~\ref{tab:aggregate} summarizes the overall performance in all propellers and datasets.

\begin{table}[h]
\caption{Propeller frequency estimation performance}
\label{tab:aggregate}
\begin{center}
\begin{tabular}{lcccc}
\hline
\textbf{Method} & \textbf{\ac{MAE}} & \textbf{\ac{RMSE}} & \textbf{\ac{MAPE}} & \textbf{Corr.} \\
\hline

\ac{CC}      & 136.2$\pm$63.0 & 360.9$\pm$55.0 & 2.75\% & 0.442 \\
\ac{HDBSCAN} & 135.3$\pm$63.3 & 360.6$\pm$54.5 & 2.73\% & 0.445 \\
\hline
\end{tabular}
\end{center}
\end{table}

We found that estimating the \ac{RPM} using both the detection methods achieved nearly identical accuracy. This is indicative since detection only plays a role in identifying the region of interest for frequency computation. Thus, the choice of the propeller detection method has negligible effect on \ac{RPM} estimation accuracy. The connected component labeling method was used in the state estimation pipeline, since it was faster than the density based clustering approach. The detected propeller regions on an event frame and their respective tracking IDs are shown in Fig.~\ref{fig:prop_id}.

\subsection{Error Analysis}
We performed an analysis  with the absolute error distributions per-propeller aggregated across all five test flight sequences for both the detection methods. The results were nearly identical for both methods exhibiting similar distribution shapes, suggesting the propeller detection method did not influence \ac{RPM} estimation. The error histograms for connected component based detection is shown in Fig.~\ref{fig:error_dist}. The histogram aggregates all individual sample-level errors with a density estimation overlay, and the vertical lines indicate mean and median error. The error for Prop~2 shows the widest spread due to an outlier dataset.

 The temporal evolution of mean absolute error with $\pm 1\sigma$ bands (first 30 seconds) is shown in Fig.~\ref{fig:error_dist}. We define the windowed error for propeller $p$ in time bin $W_k = [k, k+1)$:
\begin{equation}
    \bar{e}_p(k) = \frac{1}{|W_k|} \sum_{t \in W_k} |\Omega_{\text{est},p}(t) - \Omega_{\text{gt},p}(t)|.
\end{equation}
The error bands are computed across the datasets in each time bin. It was found that both detection methods exhibit a stable error throughout the sequence with no significant transients.

\subsection{Per-Propeller Analysis}

The redundant events of the background can cause significant noise despite the use of a spatio-temporal clustering filter. The \ac{MAE} per motor and the correlation for both methods are in Table~\ref{tab:per_prop}. Prop~1 achieves the lowest \ac{MAE}, while Prop~2 exhibits the highest. This variation is due to partial occlusions during specific flight maneuvers. Prop~4 achieves the highest correlation for both methods.

\begin{table}[h]
\caption{Per-Propeller \ac{MAE} (\ac{RPM}) and Correlation}
\label{tab:per_prop}
\begin{center}
\begin{tabular}{lcccc}
\hline
 & \textbf{Prop 1} & \textbf{Prop 2} & \textbf{Prop 3} & \textbf{Prop 4} \\
\hline

\ac{CC} \ac{MAE}       & 117.6 & 171.7 & 127.2 & 128.2 \\
\ac{HDBSCAN} \ac{MAE}  & 117.1 & 172.3 & 122.8 & 129.1 \\
\hline
\ac{CC} Corr.      & 0.372 & 0.281 & 0.461 & 0.655 \\
\ac{HDBSCAN} Corr. & 0.373 & 0.281 & 0.471 & 0.655 \\
\hline
\end{tabular}
\end{center}
\end{table}

\subsection{State Estimation Results}

To analyze the state estimation pipeline, we computed all the components of position and velocity. The orientation of the body $z$-axis was decomposed in roll and pitch angles. The comparison is reported in Table~\ref{tab:state_rmse} across the five flight sequences. The ground-truth states were obtained using \ac{RTK-GNSS} and \ac{IMU}. Since the \ac{RPM} \ac{KF} is initialized from the first measurement to avoid bias from a fixed prior, the first \SI{300}{\milli\second} of each sequence is excluded from evaluation to allow the sliding-window buffer to fill. 


\begin{table*}[htbp]
\centering
\caption{State estimation \ac{RMSE} across five test sequences. }
\label{tab:state_rmse}
\begin{tabular}{l rrr rrr rr}
\toprule
 & \multicolumn{3}{c}{\textbf{Position}} & \multicolumn{3}{c}{\textbf{Velocity}} & \multicolumn{2}{c}{\textbf{Orientation}} \\
\cmidrule(lr){2-4} \cmidrule(lr){5-7} \cmidrule(lr){8-9}
Sequence & $x$ (\si{\meter}) & $y$ (\si{\meter}) & $z$ (\si{\meter}) & $x$ (\si{\meter\per\second}) & $y$ (\si{\meter\per\second}) & $z$ (\si{\meter\per\second}) & roll (\si{\degree}) & pitch (\si{\degree}) \\
\midrule
103545 & 0.363 & 0.138 & 0.919 & 0.675 & 0.187 & 0.590 & 5.12 & 9.12 \\
103946 & 0.059 & 0.097 & 1.092 & 0.086 & 0.134 & 0.258 & 5.78 & 7.47 \\
104119 & 0.945 & 0.253 & 1.147 & 0.916 & 0.228 & 0.697 & 6.13 & 13.45 \\
104755 & 0.044 & 0.027 & 0.798 & 0.096 & 0.075 & 0.312 & 3.16 & 6.98 \\
110256 & 0.070 & 0.102 & 0.526 & 0.237 & 0.330 & 0.428 & 8.43 & 8.85 \\
\midrule
Mean$\pm$Std & 0.30$\pm$0.35 & 0.12$\pm$0.07 & 0.90$\pm$0.22 & 0.40$\pm$0.34 & 0.19$\pm$0.09 & 0.46$\pm$0.17 & 5.72$\pm$1.70 & 9.18$\pm$2.29 \\
\bottomrule
\end{tabular}
\end{table*}

The $z$-axis dominates position error due to noise from the monocular depth estimation, while the lateral axes benefit from direct camera measurements. Since the diagonal pixel length between the propellers is estimated using a noisy blob centroid, there is a systematic depth uncertainty. The estimation pipeline infers the velocity via the position-velocity cross-covariance, yielding mean \ac{RMSE} below \SI{0.5}{\meter\per\second} on all axes

\begin{figure*}[!t]
\centering
\includegraphics[width=\textwidth]{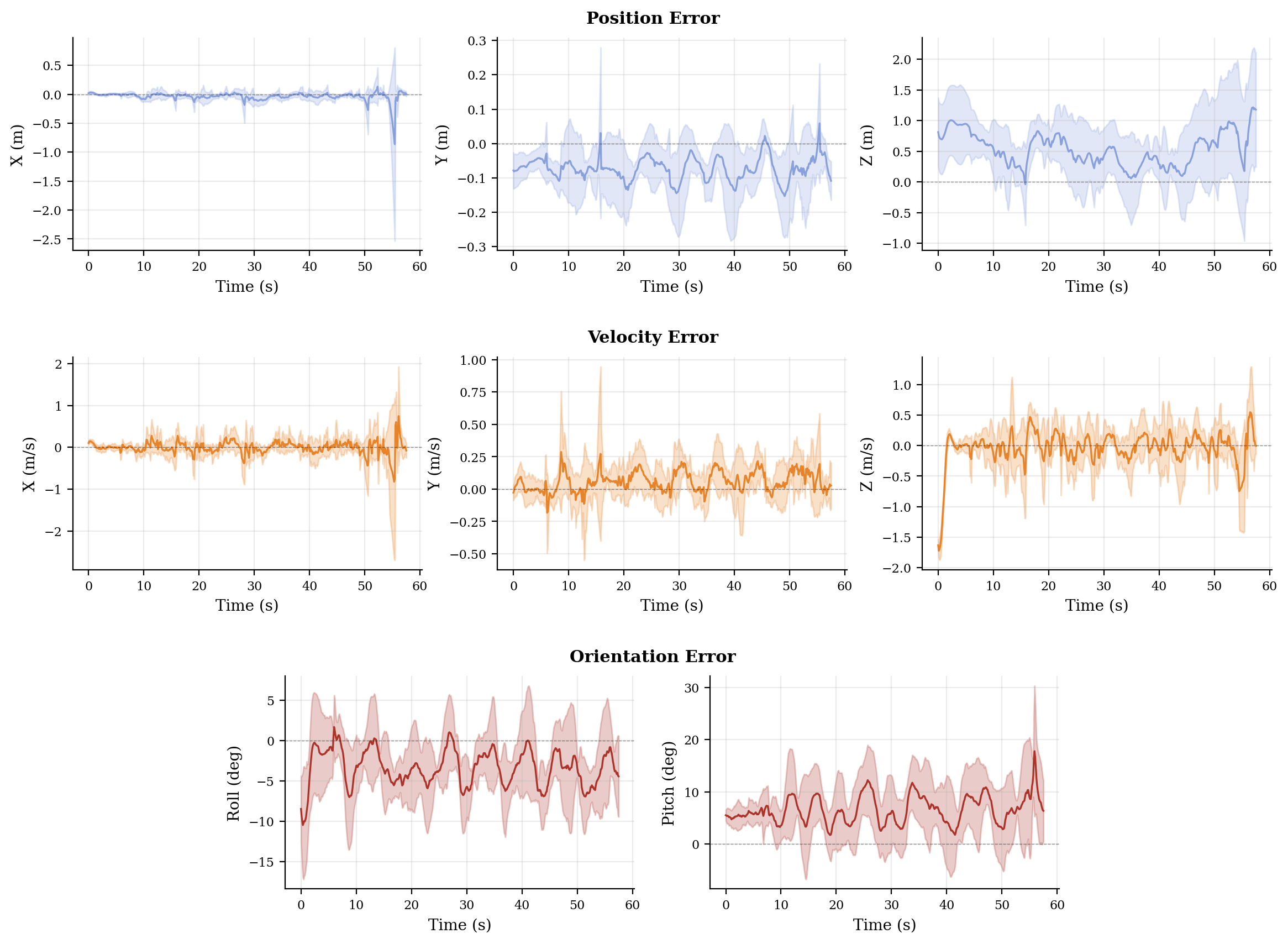}
\caption{Estimation error (mean $\pm 1\sigma$ across five sequences) over time. Top row: position error (\si{\meter}). Middle row: velocity error (\si{\meter\per\second}). Bottom row: orientation error (\si{\degree}). The shaded band represents $\pm 1$ standard deviation across sequences.}
\label{fig:error_bands}
\end{figure*}

In the five test flight sequences, the quadrotor exhibits lateral motion more than forward-backward motion. Thus, the roll range is larger than the pitch. Though the mean \ac{RMSE} for both roll and pitch are comparable, The orientation \ac{KF} estimates the roll stronger than the pitch taking advantage of left-right differential in propeller frequencies. The yaw is not estimated since it is not observable from ellipse fitting. The sequence 104119, in Table~\ref{tab:state_rmse}, exhibits the highest errors in both position and orientation, attributable to aggressive lateral maneuvers during that flight. The temporal evolution of the estimation error for all the element of state vector is shown in Fig.~\ref{fig:error_bands}.

\subsection{Discussion}
We made a few observations from the multi-dataset evaluation. The propeller detection method has a negligible effect on the accuracy of the \ac{RPM}. \ac{CC} labeling and \ac{HDBSCAN} clustering achieve nearly identical overall accuracy. The per-propeller error analysis suggests that the \ac{RPM} estimation pipeline dominates accuracy, not the spatial detection method, since the propeller region is used only to identify the region of interest. \ac{CC} processes on average faster per sequence than \ac{HDBSCAN}. Therefore, the connected component method is preferred for real-time or resource-constrained applications.

The variance of the per-propeller \ac{RPM} error indicates that propeller-specific factors such as viewing angle, blade geometry, and occlusion affect the quality of the estimation more than the detection algorithm. The accuracy of the \ac{RPM} depends on the quality of the event signal from the propeller region. The detection method may merge adjacent blobs, or miss them when the propellers are outside the frame, specifically during fast lateral motion. 


From \ac{RPM}-driven state estimation we found that the $z$-axis dominates the position error due to the monocular depth uncertainty, while lateral axes benefit from direct camera measurements. The orientation \ac{KF} recovers roll more accurately than pitch, reflecting the stronger observability of the left--right motor differential compared to the front--rear differential, due to a left-right velocity bias of the target in the test dataset. The estimated body $z$-axis from the orientation filter determines the thrust direction used in the position prediction step, enabling tilt-corrected acceleration estimates.

\section{CONCLUSIONS AND FUTURE WORK}

This paper demonstrates an event--based approach to relative state estimation for quadrotors. Propeller frequencies are extracted from the periodic event signal in each propeller's region of interest and used to enhance the orientation estimation of observed \ac{UAV}. A system of coupled \acp{KF}, one for position and the other for orientation, fuse \ac{RPM}-derived thrust and angular velocity with camera-derived position and tilt measurements. The position \ac{KF} estimates position and velocity of the quadrotor, while the orientation \ac{KF} recovers roll and pitch. The propeller frequencies provide a direct vertical acceleration signal complementary to the camera-derived position, and the differential motor \acp{RPM} enable attitude estimation through the propeller disc geometry. This proposed relative state estimation method is evaluated across five outdoor flight sequences. Future work will explore multi-quadrotor scenarios and further leverage the use of event vision to extract features based on periodic events and geometric primitives to estimate relative states.

\section*{APPENDIX}

\subsection{RPM Estimation Pipeline}
\begin{figure*}[htbp]
    \centering
    \subfloat[Propeller 1]{%
        \includegraphics[width=0.48\textwidth]{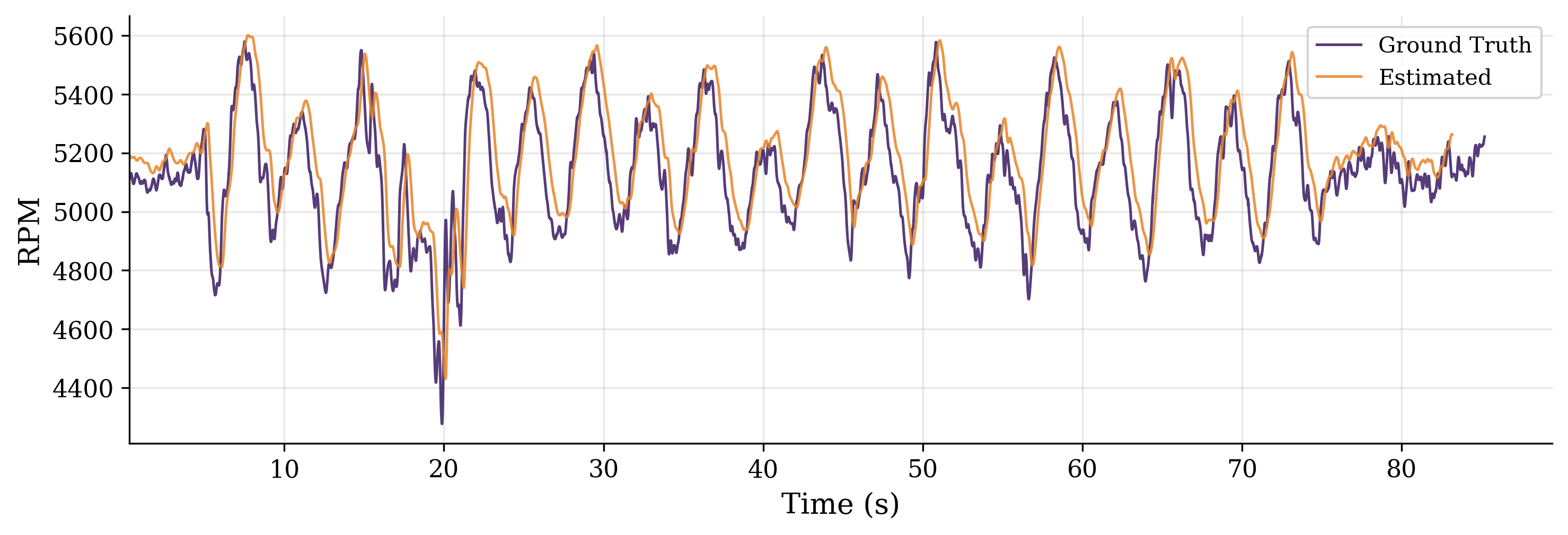}
        \label{fig:cc_prop1}
    }
    \hfill
    \subfloat[Propeller 2]{%
        \includegraphics[width=0.48\textwidth]{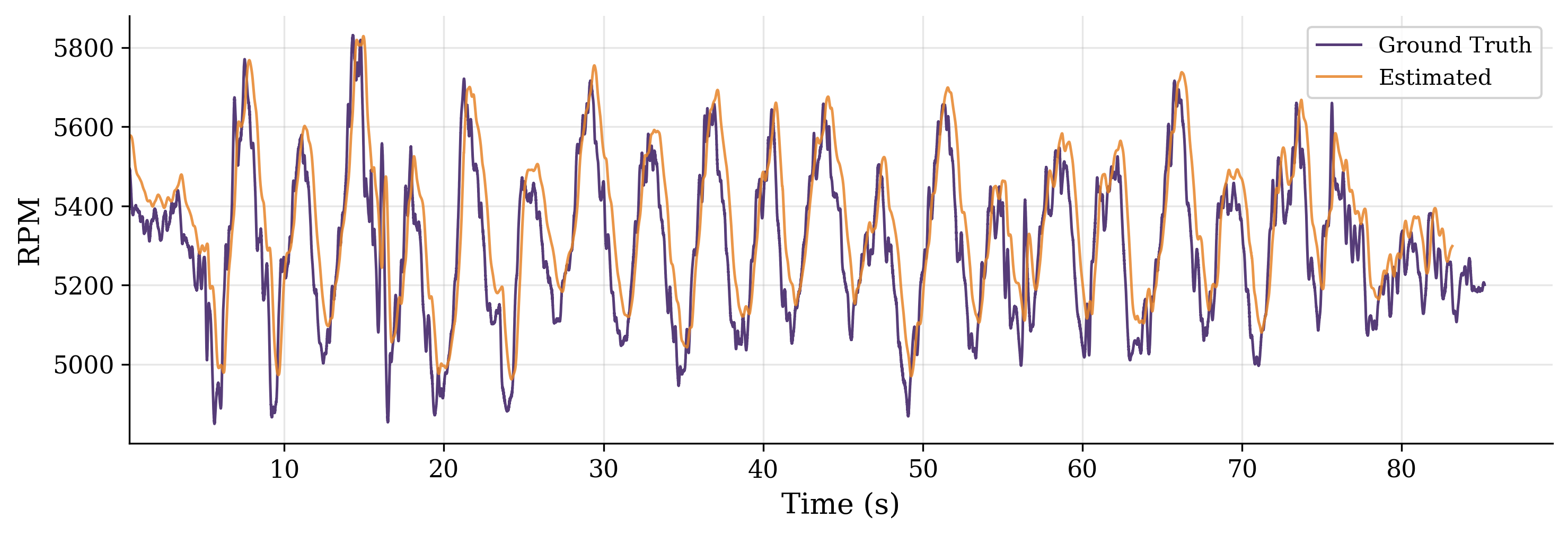}
        \label{fig:cc_prop2}
    }

    \vspace{0.5em}

    \subfloat[Propeller 3]{%
        \includegraphics[width=0.48\textwidth]{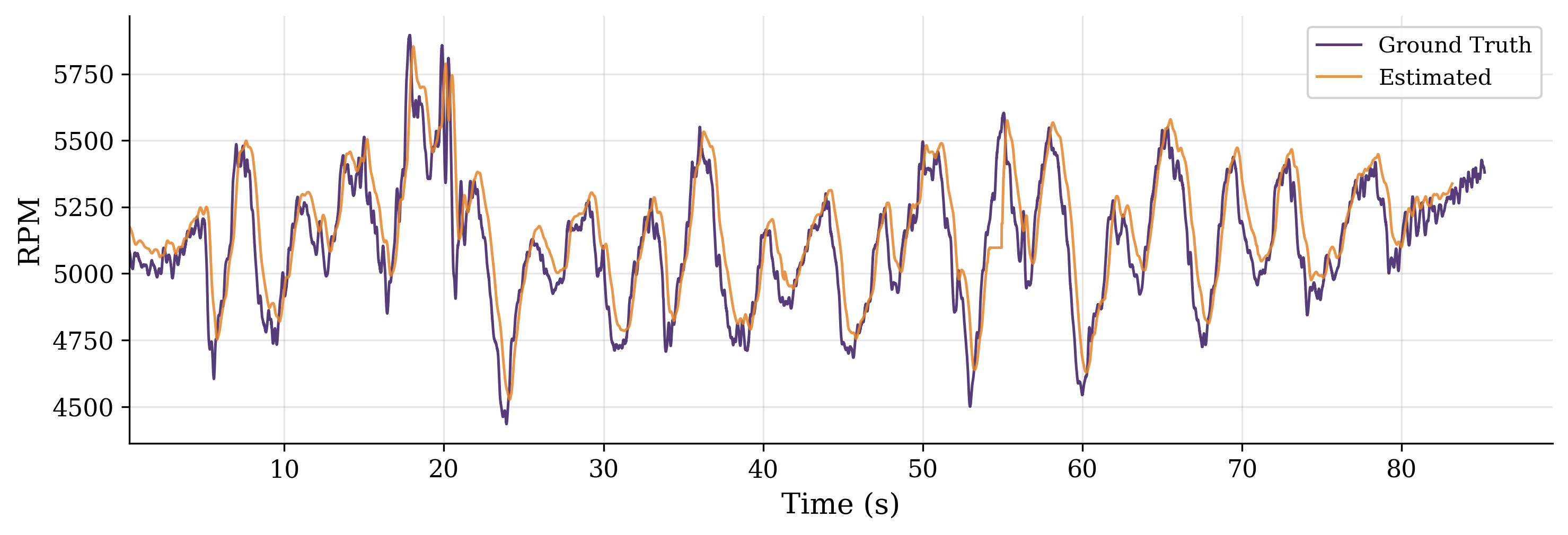}
        \label{fig:cc_prop3}
    }
    \hfill
    \subfloat[Propeller 4]{%
        \includegraphics[width=0.48\textwidth]{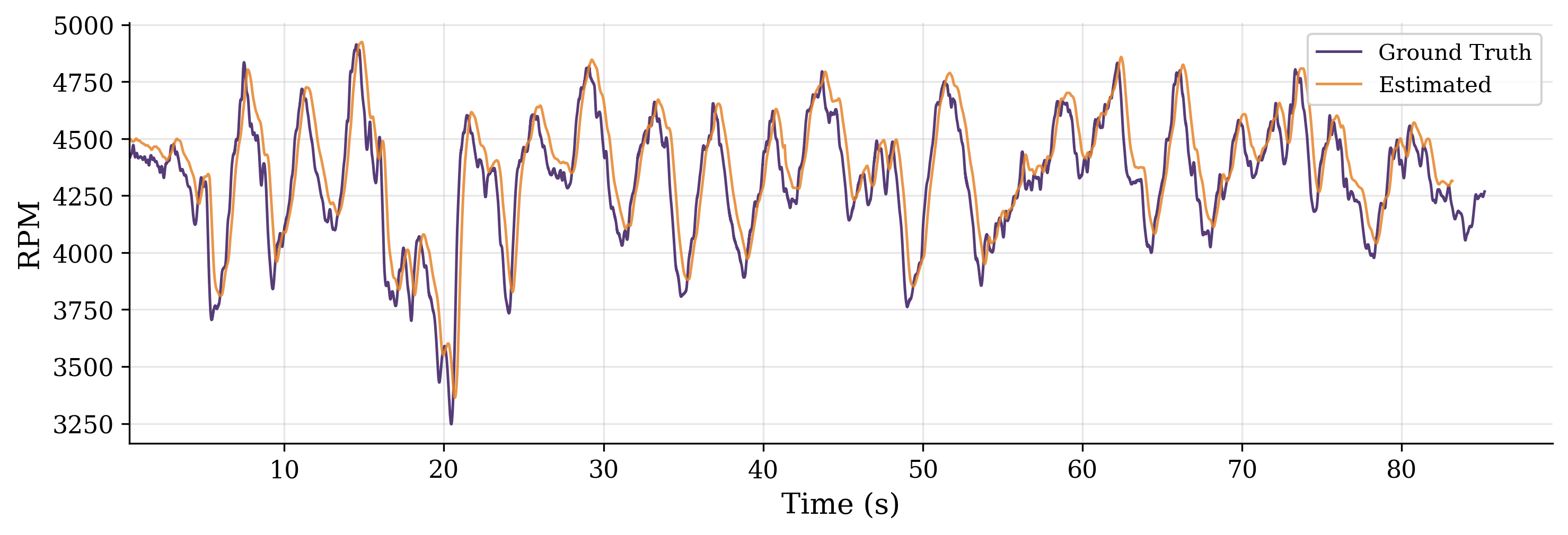}
        \label{fig:cc_prop4}
    }

    \caption{RPM estimation (red) vs ground truth (blue) for the validation sequence using connected component labeling based propeller detection. All four propellers are tracked continuously.}
    \label{fig:validation_cc}
\end{figure*}

Fig.~\ref{fig:validation_cc} shows estimated RPM versus the ground truth for a representative validation sequence (105115) using \ac{CC} detection. The estimated RPM closely tracks the ground truth across all four propellers, with deviations primarily occurring during rapid RPM transients.

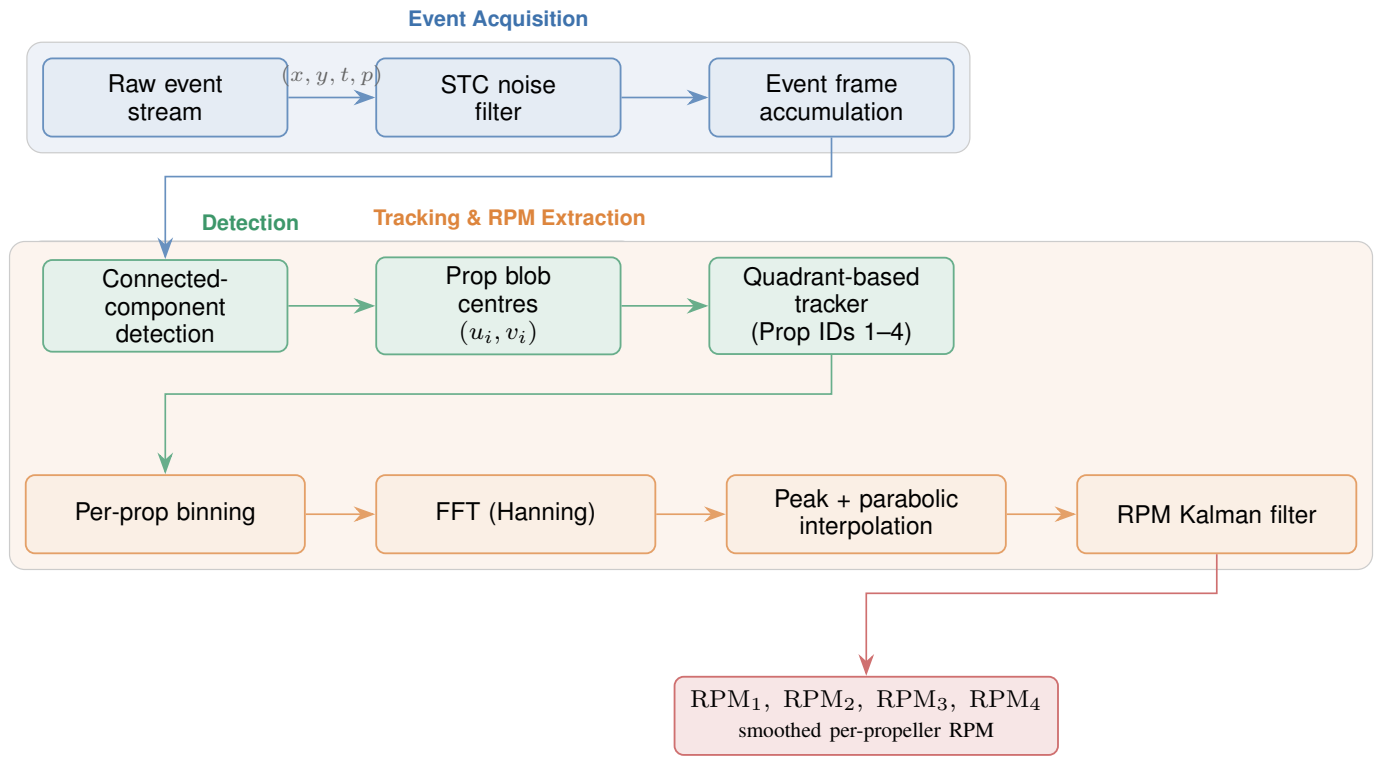
\begin{figure*}[!ht]
\centering
\resizebox{\textwidth}{!}{%
\begin{tikzpicture}[node distance=6mm and 8mm]

  \node[eventblock, text width=26mm] (raw)
    {Raw event\\stream};
  \node[eventblock, right=10mm of raw, text width=26mm] (stc)
    {STC noise\\filter};
  \node[eventblock, right=10mm of stc, text width=26mm] (frame)
    {Event frame\\accumulation};

  \draw[myarrow=eventblue!70] (raw) -- (stc)
    node[midway, above, font=\scriptsize\sffamily, text=black!60]
    {$(x,y,t,p)$};
  \draw[myarrow=eventblue!70] (stc) -- (frame);

  \begin{scope}[on background layer]
    \node[stagefit=eventblue!8,
          fit=(raw)(stc)(frame),
          label={[stagelabel=eventblue!90]above:
                 Event Acquisition}] {};
  \end{scope}

  \node[procblock, below=14mm of raw, text width=26mm] (cc)
    {Connected-\\component\\detection};
  \node[procblock, right=10mm of cc, text width=26mm] (blobs)
    {Prop blob\\centres\\$(u_i, v_i)$};

  \draw[myarrow=eventblue!70] (frame.south) -- ++(0,-4.5mm) -| (cc.north);
  \draw[myarrow=procgreen!70] (cc) -- (blobs);

  \begin{scope}[on background layer]
    \node[stagefit=procgreen!8,
          fit=(cc)(blobs),
          label={[stagelabel=procgreen!90, xshift=5mm]above left:
                 Detection}] {};
  \end{scope}

  \node[procblock, right=10mm of blobs, text width=26mm] (track)
    {Quadrant-based\\tracker\\(Prop IDs 1--4)};
  \node[measblock, below=14mm of cc, text width=30mm] (bin)
    {Per-prop binning\\[-2pt]};
  \node[measblock, right=8mm of bin, text width=30mm] (acorr)
    {FFT (Hanning)\\[-2pt]};

  \node[measblock, right=8mm of acorr, text width=30mm] (peak)
    {Peak + parabolic\\interpolation\\[-2pt]};
  \node[measblock, right=8mm of peak, text width=30mm] (kfrpm)
    {RPM Kalman filter\\[-2pt]};

  \draw[myarrow=procgreen!70] (blobs) -- (track);
  \draw[myarrow=procgreen!70] (track.south) -- ++(0,-4.5mm) -| (bin.north);
  \draw[myarrow=measorange!70] (bin) -- (acorr);
  \draw[myarrow=measorange!70] (acorr) -- (peak);
  \draw[myarrow=measorange!70] (peak) -- (kfrpm);

  \begin{scope}[on background layer]
    \node[stagefit=measorange!8,
          fit=(track)(bin)(acorr)(peak)(kfrpm),
          label={[stagelabel=measorange!90, xshift=15mm]above left:
                 Tracking \& RPM Extraction}] {};
  \end{scope}

  \node[outblock, below=14mm of peak, text width=42mm] (out)
    {$\mathrm{RPM}_1,\;\mathrm{RPM}_2,\;\mathrm{RPM}_3,\;\mathrm{RPM}_4$\\[1pt]
     {\normalfont\scriptsize smoothed per-propeller RPM }};

  \draw[myarrow=outputred!70] (kfrpm.south) -- ++(0,-4.5mm) -| (out.north);


\end{tikzpicture}}
\caption{The propeller RPM estimation pipeline.  Raw events from the event camera are filtered, accumulated into frames, and processed through connected component detection to locate spinning propeller blobs.  A tracker assigns stable Prop~IDs, after which per-propeller FFT extracts the blade-passage frequency.}
\label{fig:rpm_flow}
\end{figure*}

Figure~\ref{fig:rpm_flow} shows the processing stages for propeller frequency extraction from event camera data.

\bibliographystyle{IEEEtran}   
\bibliography{ref}

\end{document}